\documentclass{article}
\usepackage{amsmath, amssymb, amsthm} 
\usepackage{bm} 
\usepackage{graphicx} 
\usepackage{booktabs} 
\usepackage{enumitem} 
\usepackage{algorithm} 
\usepackage[noend]{algpseudocode} 
\usepackage[a4paper, margin=1in]{geometry} 
\usepackage{mathtools} 
\usepackage{cite,url} 
\usepackage{fancyhdr}

\newcommand{\vb}[1]{\boldsymbol{#1}}

\DeclareMathOperator{\Std}{Std} 
\DeclareMathOperator{\clip}{clip}
\newcommand{\E}{\mathbb{E}}

\newcommand{\mean}[1]{\mathbb{E}\left[#1\right]}

\begin{document}

\title{\bfseries AM-PPO: (Advantage) Alpha-Modulation with Proximal Policy Optimization}
\author{Soham Sane \\ \small Collins Aerospace (RTX)}
\date{\today}
\maketitle

\hrule height 2pt
\begin{abstract}
Proximal Policy Optimization (PPO) is a widely used reinforcement learning algorithm that heavily relies on accurate advantage estimates for stable and efficient training. However, raw advantage signals can exhibit significant variance, noise, and scale-related issues, impeding optimal learning performance. To address this challenge, we introduce Advantage Modulation PPO (AM-PPO), a novel enhancement of PPO that adaptively modulates advantage estimates using a dynamic, non-linear scaling mechanism. This adaptive modulation employs an alpha controller that dynamically adjusts the scaling factor based on evolving statistical properties of the advantage signals, such as their norm, variance, and a predefined target saturation level. By incorporating a \(\tanh\)-based gating function driven by these adaptively scaled advantages, AM-PPO reshapes the advantage signals to stabilize gradient updates and improve the conditioning of the policy gradient landscape. Crucially, this modulation also influences value function training by providing consistent and adaptively conditioned learning targets. Empirical evaluations across standard continuous control benchmarks demonstrate that AM-PPO achieves superior reward trajectories, exhibits sustained learning progression, and significantly reduces the clipping required by adaptive optimizers. These findings underscore the potential of advantage modulation as a broadly applicable technique for enhancing reinforcement learning optimization.
\end{abstract}
\vspace{7pt}
\hrule height 1pt

\section{Introduction}
Policy gradient methods are a cornerstone of modern reinforcement learning (RL), with Proximal Policy Optimization (PPO) \cite{Schulman2017ProximalPO} being a particularly successful and widely adopted actor-critic algorithm. PPO learns a policy \(\pi_\theta(a|s)\) and a state-value function \(V_\phi(s)\), leveraging advantage estimates \(A^{\pi}(s,a)\) to guide policy updates. While techniques like Generalized Advantage Estimation (GAE) \cite{Schulman2015HighDimensionalCC} are instrumental in reducing the variance of these estimates, the resulting learning signal can still exhibit suboptimal scaling or residual noise, potentially hindering learning progress or leading to inefficiencies.

Inspired by adaptive optimization methods such as Dynamic AlphaGrad (DynAG) \cite{DynAG2025}, which dynamically adjust learning signals based on their statistical properties, we propose AM-PPO (Alpha Modulation with PPO). AM-PPO enhances the standard PPO framework by introducing a novel mechanism to modulate raw advantage estimates prior to their use in the policy and value function learning updates. This modulation is governed by an adaptive controller whose internal state—most notably an adaptive scaling factor \(\alpha_A\)—evolves based on observed statistics of the advantages, such as their \(L_2\) norm, standard deviation, and a target saturation level for a non-linearly transformed version of these advantages.

The central hypothesis of this work is that by dynamically rescaling and reshaping the advantage signal, AM-PPO can foster more stable and efficient policy learning. This is achieved by conditioning the policy gradient more effectively, potentially leading to improved optimization landscapes and more consistent learning trajectories. AM-PPO employs a specific modulation formula: it scales the absolute value of the raw advantages by a gating term. This gate is constructed from a shared scaling factor \(\kappa_{\text{shared}}\) and a \(\tanh\) function applied to the adaptively scaled, normalized advantages. Crucially, the state-value function \(V_\phi(s)\) in AM-PPO is also trained using targets derived from these modulated advantages, ensuring that both the actor and critic learn from a consistently transformed signal. This paper details the AM-PPO framework, elucidates its adaptive control mechanisms, discusses the theoretical and practical implications of its design, and presents an empirical evaluation of its performance characteristics.

\newpage

\section{PPO with Group Advantage Modulation (AM-PPO)}
\label{sec:AM-PPO-full}

AM-PPO retains the core structure of PPO but modifies how advantage estimates influence both the policy and value function updates. It employs a standard actor-critic setup and GAE for raw advantage calculation but introduces an adaptive modulation layer for these advantages, which are then used for both policy and value learning.

\subsection{Preliminaries: PPO and Advantage Estimation}
The agent learns a policy \(\pi_\theta(a|s)\) and a state-value function \(V_\phi(s)\). Data is collected by running the policy for \(N_{\text{steps}}\) in \(N_{\text{envs}}\) parallel environments. Raw advantage estimates \(A^{\text{raw}}_t\) are computed using GAE:
\begin{equation}
    A^{\text{raw}}_t = \sum_{l=0}^{N_{\text{steps}}-t-1} (\gamma \lambda)^l \delta_{t+l}
    \label{eq:gam_gae}
\end{equation}
where \(\gamma\) is the discount factor, \(\lambda\) is the GAE smoothing parameter, and \(\delta_t\) is the TD error:
\begin{equation}
    \delta_t = r_t + \gamma V_\phi(s_{t+1})(1-d_t) - V_\phi(s_t)
    \label{eq:gam_td_error}
\end{equation}
Here, \(r_t\) is the reward at timestep \(t\), and \(d_t\) indicates if \(s_{t+1}\) is terminal. In standard PPO, the value function \(V_\phi\) is trained towards the on-policy target \(A^{\text{raw}}_t + V_\phi(s_t)\). AM-PPO modifies this target as described later.

\subsection{Adaptive Advantage Modulation Mechanism}
\label{subsec:gam_modulation_mechanism}
After calculating the raw GAE advantages \(\vb{A}_{\text{raw}}\) for a batch of experience, AM-PPO introduces an adaptive modulation step. This mechanism is governed by a controller with persistent Exponential Moving Average (EMA) states: \(\alpha_{A,\text{ema}}\) for an adaptive scaling factor, and \(s_{\text{prev},A,\text{ema}}\) for the observed saturation ratio of a transformed advantage signal. These EMA states are updated once per iteration based on the statistics of the entire batch of \(\vb{A}_{\text{raw}}\).

During the PPO update epochs, for each minibatch \(\vb{A}_{\text{raw}, mb}\), the modulation is applied using the \textbf{frozen} values of \(\alpha_{A,\text{ema}}\) and \(s_{\text{prev},A,\text{ema}}\) (i.e., their values at the start of the current iteration's update phase). The transformation proceeds as follows:

\begin{enumerate}
    \item \textbf{Minibatch Statistics:} Compute the \(L_2\) norm \(N_{A,mb} = \lVert \vb{A}_{\text{raw}, mb} \rVert_2\) and standard deviation \(\sigma_{A,mb} = \Std(\vb{A}_{\text{raw}, mb}) + \epsilon_A\). If \(N_{A,mb} < \epsilon_A\), \(\vb{A}_{\text{mod}, mb} = \vb{A}_{\text{raw}, mb}\) and subsequent steps are skipped.
    
    \item \textbf{Read Frozen Adaptive Scaler:} Let \(\alpha_{A,\text{current}} = \alpha_{A,\text{ema}}\) (the frozen value for this iteration's updates).
    
    \item \textbf{Normalize and Scale for \(\tanh\) Input:} Normalize the raw advantages by their \(L_2\) norm and scale by \(\alpha_{A,\text{current}}\) to get \(\vb{Z}_{A,mb}\):
    \begin{align}
        \tilde{\vb{A}}_{\text{raw}, mb} &= \frac{\vb{A}_{\text{raw}, mb}}{N_{A,mb} + \epsilon_A} \label{eq:gam_adv_norm_l2} \\
        \vb{Z}_{A,mb} &= \alpha_{A,\text{current}} \cdot \tilde{\vb{A}}_{\text{raw}, mb} \label{eq:gam_Z_A}
    \end{align}
    
    \item \textbf{Apply Group Modulation Formula:} The final modulated advantages \(\vb{A}_{\text{mod}, mb}\) are computed using a shared scaling factor \(\kappa_{\text{shared}}\)). The absolute value of the raw advantages is multiplied by a gating term:
    \begin{equation}
        \vb{A}_{\text{mod}, mb} = |\vb{A}_{\text{raw}, mb}| \odot \left( \kappa_{\text{shared}} \cdot \tanh(\vb{Z}_{A,mb}) \right)
        \label{eq:gam_A_mod_final}
    \end{equation}
    The term \( M_{\text{gate}} = (\kappa_{\text{shared}} \cdot \tanh(\vb{Z}_{A,mb})\) acts as a gate. The sign of the modulated advantage \(\vb{A}_{\text{mod}, mb}\) is determined by the sign of this gate \(M_{\text{gate}}\), while its magnitude is a product of \(|\vb{A}_{\text{raw}, mb}|\) and \(|M_{\text{gate}}|\).
\end{enumerate}

\paragraph{EMA State Updates (Once per Iteration):}
The persistent EMA states \(\alpha_{A,\text{ema}}\) and \(s_{\text{prev},A,\text{ema}}\) are updated once per training iteration, typically before the PPO update epochs begin, by applying the full controller logic to the entire batch of raw advantages \(\vb{A}_{\text{raw}}\) (or a large representative sample). This full logic includes:
\begin{enumerate}
    \item Computing batch-wide \(N_A\) and \(\sigma_A\).
    \item Calculating a target alpha \(\hat{\alpha}_A\) using the controller's shared scaling factor \(\kappa_{\text{shared}}\) (identical to the one in Eq. \ref{eq:gam_A_mod_final}), the batch norm/std ratio, and saturation feedback:
    \begin{equation}
        \hat{\alpha}_A = \kappa_{\text{shared}} \cdot \frac{N_{A} + \epsilon_A}{\sigma_{A}} \cdot \left(\frac{p_{\star,A}}{s_{\text{prev},A,\text{ema}} + \epsilon_A}\right)^{\eta_A}
        \label{eq:gam_hat_alpha_A}
    \end{equation}
    \item Updating \(\alpha_{A,\text{ema}}\) towards \(\hat{\alpha}_A\) via EMA with smoothing factor \(\rho_A\), and clamping it:
    \begin{equation}
        \alpha_{A,\text{ema}} \leftarrow \clip\big((1 - \rho_A) \alpha_{A,\text{ema}} + \rho_A \hat{\alpha}_A, \alpha_{\min,A}, \alpha_{\max,A}\big)
        \label{eq:gam_alpha_A_ema_update}
    \end{equation}
    \item Computing \(\vb{Z}_A\) for the batch using the newly updated \(\alpha_{A,\text{ema}}\) (as \(\alpha_{A,\text{current}}\) for this EMA update step).
    \item Observing the current saturation for the batch \(s_{\text{curr},A} = \mean(\lvert \vb{Z}_{A} \rvert > \tau_A)\).
    \item Updating \(s_{\text{prev},A,\text{ema}}\) towards \(s_{\text{curr},A}\) via EMA with smoothing factor \(\rho_{\text{sat},A}\):
    \begin{equation}
        s_{\text{prev},A,\text{ema}} \leftarrow (1 - \rho_{\text{sat},A}) s_{\text{prev},A,\text{ema}} + \rho_{\text{sat},A} s_{\text{curr},A}
        \label{eq:gam_sat_ema_update}
    \end{equation}
\end{enumerate}

\subsection{PPO Learning Update}
The PPO learning update proceeds for \(N_{\text{epochs}}\) over minibatches.
\begin{enumerate}[label=(\roman*)]
    \item \textbf{Normalize Modulated Advantages:} If enabled by a configuration setting, the \(\vb{A}_{\text{mod}, mb}\) from Equation \ref{eq:gam_A_mod_final} is further normalized to have mean zero and unit variance, yielding \(\hat{\vb{A}}_{\text{mod}, mb}\). Otherwise, \(\hat{\vb{A}}_{\text{mod}, mb} = \vb{A}_{\text{mod}, mb}\).

    \item \textbf{Policy Loss (\(\mathcal{L}^{\pi}\)):} The policy parameters \(\theta\) are updated by maximizing the PPO clipped surrogate objective using \(\hat{\vb{A}}_{\text{mod}, mb}\). Let \(r_j(\theta) = \frac{\pi_\theta(a_j|s_j)}{\pi_{\theta_{\text{old}}}(a_j|s_j)}\) be the probability ratio.
    \begin{equation}
        \mathcal{L}^{\pi}(\theta) = \E_{j \in mb} \left[ \min \left( r_j(\theta) \hat{A}_{\text{mod}, j}, \clip(r_j(\theta), 1-\epsilon_{\text{clip}}, 1+\epsilon_{\text{clip}}) \hat{A}_{\text{mod}, j} \right) \right]
        \label{eq:gam_ppo_policy_loss}
    \end{equation}

    \item \textbf{Value Loss (\(\mathcal{L}^V\)):} The value function parameters \(\phi\) are updated by minimizing the error between \(V_\phi(s_j)\) and the target \(V_{\text{target}, j}\). This target is constructed using the \textbf{modulated advantages} \(A_{\text{mod}, j}\) (from Eq. \ref{eq:gam_A_mod_final}, prior to optional normalization for the policy loss) and the value estimates from before the update epoch, \(V_{\phi_{\text{old}}}(s_j)\):
    \begin{equation}
        V_{\text{target}, j} = A_{\text{mod}, j} + V_{\phi_{\text{old}}}(s_j)
        \label{eq:gam_value_target_ppo_gam}
    \end{equation}
    The value loss is then:
    \begin{align}
    \mathcal{L}^V(\phi) = \E_{j \in \text{mb}} \Big[ \max \big( 
        & \left( V_\phi(s_j) - V_{\text{target}, j} \right)^2, \nonumber \\
        & \left( V_{\phi_{\text{old}}}(s_j) 
        + \clip\left(V_\phi(s_j) - V_{\phi_{\text{old}}}(s_j), 
        -\epsilon_{\text{clip}}, \epsilon_{\text{clip}}\right) 
        - V_{\text{target}, j} \right)^2 
    \big) \Big]
    \label{eq:gam_ppo_value_loss}
    \end{align}
    
    \item \textbf{Entropy Loss (\(\mathcal{L}^S\)):} An optional entropy bonus \( - \beta_{\text{ent}} \E [ H(\pi_\theta(\cdot|s_j)) ] \) encourages exploration.
    
    \item \textbf{Total Loss and Parameter Update:} The combined loss \(\mathcal{L}(\theta, \phi) = -\mathcal{L}^{\pi}(\theta) + c_V \mathcal{L}^{V}(\phi) + \mathcal{L}^{S}(\theta)\) is optimized.
\end{enumerate}

\subsection{Hyperparameters}
Key hyperparameters for AM-PPO are summarized in Table \ref{tab:ppogam_hyperparams}.

\begin{table}[H]
\centering
\caption{Key Hyperparameters for AM-PPO.}
\label{tab:ppogam_hyperparams}
\begin{tabular}{@{}ll@{}}
\toprule
Parameter                        & Description                                                        \\ \midrule
\multicolumn{2}{l}{\textbf{PPO Core}}                                                               \\
\(N_{\text{steps}}\)             & Number of environment steps per iteration per environment          \\
\(N_{\text{envs}}\)              & Number of parallel environments                                    \\
\(N_{\text{epochs}}\)            & Number of optimization epochs per iteration                      \\
\(N_{\text{minibatches}}\)       & Number of minibatches per epoch                                    \\
\(\gamma\)                       & Discount factor                                                    \\
\(\lambda\)                      & GAE lambda parameter                                               \\
\(\epsilon_{\text{clip}}\)       & PPO clipping coefficient                                           \\
\(c_V\)                          & Value function loss coefficient                                   \\
\(\beta_{\text{ent}}\)           & Entropy bonus coefficient                                         \\
\(\text{lr}\)                    & Learning rate                                                      \\
\(\text{clip\_vloss}\)           & Boolean: enable/disable value loss clipping                       \\
\(\text{norm\_adv}\)             & Boolean: enable/disable normalization of final policy advantages  \\
\(\text{max\_grad\_norm}\)       & Gradient clipping threshold                                        \\ \midrule
\multicolumn{2}{l}{\textbf{Advantage Modulation}}                                     \\
\(\kappa_{\text{shared}}\)       & Shared scaling factor for controller's \(\hat{\alpha}_A\) (Eq. \ref{eq:gam_hat_alpha_A}) \\
                                 & and for \(\tanh(\vb{Z}_{A,mb})\) output in Eq. \ref{eq:gam_A_mod_final}. \\
\(\tau_A\)                       & Saturation threshold for controller input \(Z_A\)                   \\
\(p_{\star,A}\)                  & Target saturation probability for \(Z_A\)                            \\
\(\eta_A\)                       & Strength of saturation feedback for \(\hat{\alpha}_A\)                \\
\(\rho_A\)                       & EMA smoothing factor for \(\alpha_{A,\text{ema}}\)                   \\
\(\rho_{\text{sat},A}\)          & EMA smoothing factor for \(s_{\text{prev},A,\text{ema}}\)            \\
\(\alpha_{\min,A}\)              & Minimum clamp value for \(\alpha_{A,\text{ema}}\)                    \\
\(\alpha_{\max,A}\)              & Maximum clamp value for \(\alpha_{A,\text{ema}}\)                    \\
\(\epsilon_A\)                   & Small constant for numerical stability in controller             \\
\(\alpha_{A,\text{ema}}^{(0)}\)  & Initial value for \(\alpha_{A,\text{ema}}\) state                   \\
\(s_{\text{prev},A,\text{ema}}^{(0)}\) & Initial value for \(s_{\text{prev},A,\text{ema}}\) state           \\ \bottomrule
\end{tabular}
\end{table}

\begin{figure}[H]
    \centering
    \includegraphics[width=0.8\linewidth]{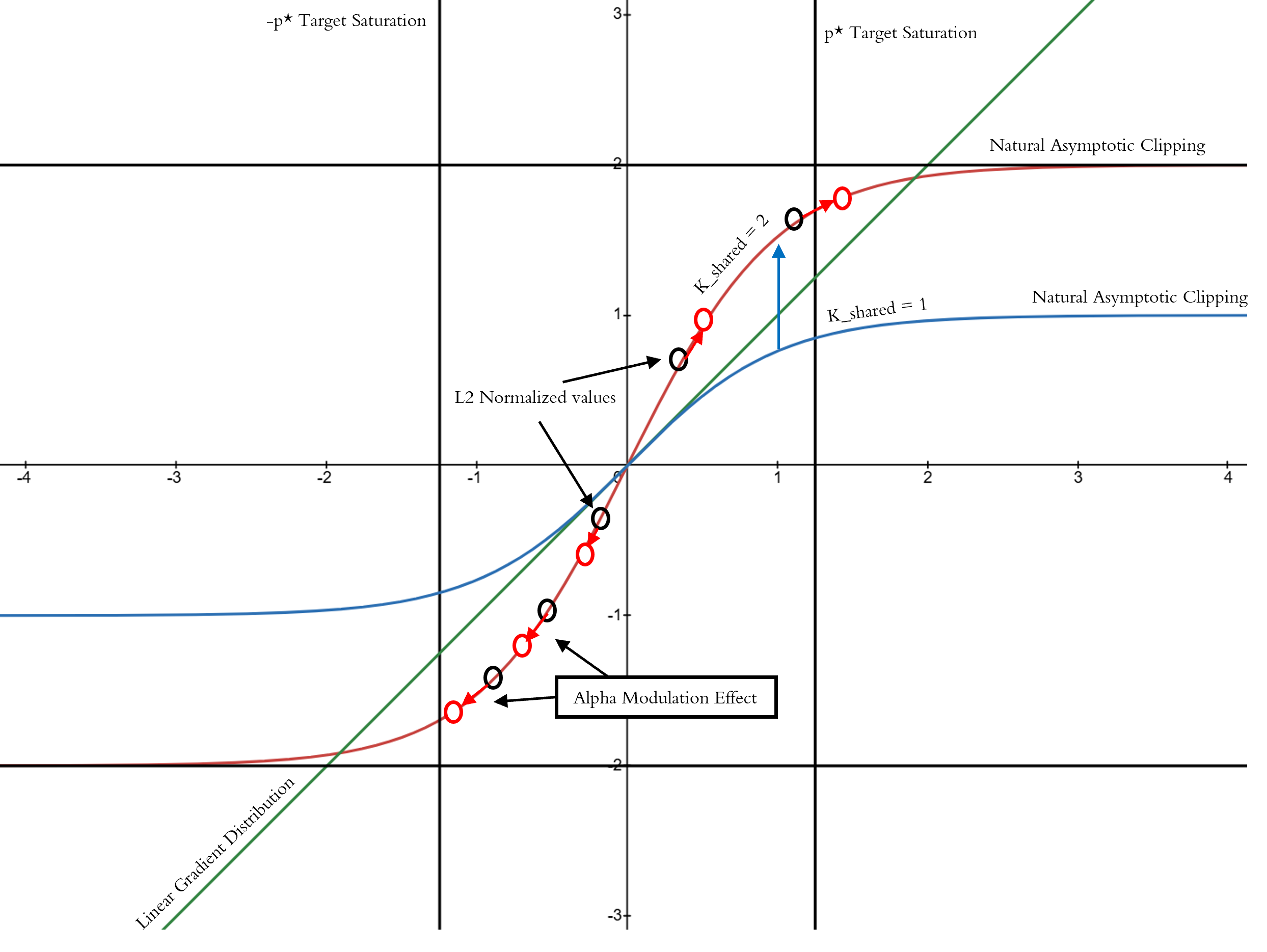}
    \caption{Alpha Modulation Effect}
    \label{fig:AlphaModulationGraphic}
\end{figure}

Figure \ref{fig:AlphaModulationGraphic} showcases the mathematical understanding and effects of the alpha modulation controller given a set of normalized values.

\section{Theoretical Motivations for Advantage Modulation}
\label{app:theoretical_motivations_appendix_e} 

The design of the AM-PPO framework, particularly its adaptive advantage modulation mechanism, is predicated on the understanding that raw advantage estimates in reinforcement learning often exhibit complex, non-linear characteristics that are not optimally addressed by simple linear transformations or static normalization schemes.

\subsection{The Nature of Advantage Signals and the Insufficiency of Linear Models}
In complex RL environments, the state-value function \(V_\phi(s)\) and consequently the advantage function \(A^{\pi}(s,a)\) are typically approximated by highly non-linear functions, such as deep neural networks (e.g., Multi-Layer Perceptrons, FFNs) \cite{Goodfellow2016DeepLearning}. The distribution of true advantages across states and actions, and thus their empirical estimates \(A^{\text{raw}}_t\), can therefore be expected to be similarly complex, multimodal, and non-Gaussian.

A purely linear transformation or a static scaling of such advantage signals is unlikely to be universally optimal across different stages of training or diverse sets of experiences. For instance, the scale and variance of relative advantages within an on-policy buffer can change drastically as the policy improves or shifts its exploration focus. Furthermore, the raw advantage distribution might contain outliers or exhibit heavy tails that can disproportionately influence the policy gradient if not appropriately managed \cite{Fujimoto2021HuberAC}. The objective of a policy gradient update is to find an ascent direction in a high-dimensional, non-convex loss landscape. A simple linear interpretation of the advantage signal (e.g., using it directly without adaptive conditioning) may fail to capture the nuanced information required for stable and efficient navigation of this landscape.

\subsection{AM-PPO as a Non-Linear, Adaptive Signal Conditioner}
AM-PPO introduces specific mechanisms designed to address the aforementioned complexities by performing adaptive, non-linear conditioning of the advantage signals.

\paragraph{The Role of the \(\tanh\) Function:}
The use of the \(\tanh\) function in the modulation formula (Eq.~\ref{eq:gam_A_mod_final}) is a deliberate choice to introduce a bounded, non-linear transformation.
\begin{enumerate}
    \item \textbf{Boundedness:} The \(\tanh\) function maps its input \(\vb{Z}_{A,mb}\) to the range \((-1, 1)\). When scaled by \(\kappa_{\text{shared}}\), this ensures that the gating term \(M_{\text{gate}}\) is bounded, preventing extremely large (potentially noisy) values in \(\vb{Z}_{A,mb}\) from exerting unbounded influence on the modulated advantages \(\vb{A}_{\text{mod}, mb}\). This provides robustness against outliers.
    \item \textbf{Non-Linearity and Saturation:} \(\tanh\) is inherently non-linear, with a sigmoidal shape. It exhibits high sensitivity to inputs around zero and progressively less sensitivity (saturation) for inputs of larger magnitude. This non-linearity allows the modulation to differentially treat advantages based on their (adaptively scaled) magnitude. Small-to-moderate scaled advantages might be transformed more linearly, while large ones are compressed. This shaping can be beneficial for stabilizing updates, ensuring that the learning signal does not become excessively large. The target saturation probability \(p_{\star,A}\) within the controller (Eq.~\ref{eq:gam_hat_alpha_A}) directly aims to manage the extent to which advantage signals fall into these saturated or sensitive regions of the \(\tanh\) curve.
\end{enumerate}

\paragraph{The Adaptive Controller for Relative Adjustment:}
The core of the AM-PPO's conditioning lies in its adaptive controller, which determines the scaling factor \(\alpha_{A,\text{current}}\) used to compute \(\vb{Z}_{A,mb}\) (Eq.~\ref{eq:gam_Z_A}). This controller enables a form of relative normalization and adjustment:
\begin{enumerate}
    \item \textbf{Normalization and Initial Scaling:} The raw advantages \(\vb{A}_{\text{raw}, mb}\) are first normalized by their \(L_2\) norm \(\tilde{\vb{A}}_{\text{raw}, mb} = \frac{\vb{A}_{\text{raw}, mb}}{N_{A,mb} + \epsilon_A}\) (Eq.~\ref{eq:gam_adv_norm_l2}). This step removes the initial batch-specific magnitude, allowing \(\alpha_{A,\text{current}}\) to subsequently impose a controlled scaling.
    \item \textbf{Dynamic Adaptation of \(\alpha_A\):} The target \(\hat{\alpha}_A\) (Eq.~\ref{eq:gam_hat_alpha_A}) is calculated based on statistics of the \textit{current batch} of advantages (\(N_A, \sigma_A\)) and feedback from the \textit{historical} saturation level (\(s_{\text{prev},A,\text{ema}}\)).
        \begin{itemize}
            \item The term \(\frac{N_A + \epsilon_A}{\sigma_A}\) captures the relationship between the overall magnitude and the dispersion of advantages. A distribution that is widespread relative to its mean magnitude might require different scaling than a more concentrated one.
            \item The term \(\left(\frac{p_{\star,A}}{s_{\text{prev},A,\text{ema}} + \epsilon_A}\right)^{\eta_A}\) provides closed-loop control. If the observed saturation \(s_{\text{prev},A,\text{ema}}\) deviates from the target \(p_{\star,A}\), \(\hat{\alpha}_A\) is adjusted to push the saturation back towards the target. This ensures that the input to the \(\tanh\) function is consistently conditioned over time, adapting to changing advantage distributions.
        \end{itemize}
    \item \textbf{EMA for Temporal Smoothing:} The EMA updates for \(\alpha_{A,\text{ema}}\) and \(s_{\text{prev},A,\text{ema}}\) (Eqs.~\ref{eq:gam_alpha_A_ema_update}, \ref{eq:gam_sat_ema_update}) ensure that these adaptive parameters evolve smoothly, incorporating information from past patterns of advantage statistics rather than overreacting to instantaneous fluctuations in a single batch. This captures the ``nuance'' of historical trends.
\end{enumerate}
The combination of these elements---L2 normalization, adaptive scaling via a feedback-controlled \(\alpha_A\), and the bounded non-linear \(\tanh\) transformation---results in a sophisticated signal processing pipeline. It does not merely apply a fixed transformation but dynamically adjusts the characteristics of the advantage signals relative to their own evolving statistical properties and predefined operational targets (i.e., target saturation).

While this paper does not present a formal mathematical proof demonstrating the universal optimality of AM-PPO's specific modulation strategy, the design is grounded in the well-understood premise that advantage signals in RL are complex and benefit from adaptive, non-linear conditioning. The components of AM-PPO are principled heuristics aimed at achieving such conditioning: the \(\tanh\) function provides controlled non-linearity and boundedness, while the adaptive alpha controller facilitates dynamic, relative adjustments based on observed signal characteristics and historical performance. The empirical results presented in the main body of this work (Section~\ref{sec:experimentation}) serve to validate the practical efficacy of these theoretically motivated design choices.

\subsection{Related Work}
\label{subsec:related_work}

The approach presented in AM-PPO intersects with several established themes in reinforcement learning, particularly concerning the processing and normalization of learning signals.

The concept of adaptive signal processing appears in methods that normalize rewards or returns. For instance, AN-SAC \cite{Liu2023SoftAC} adaptively normalizes cumulative rewards in Soft Actor-Critic to stabilize learning across environments with varying reward scales. While sharing the spirit of adaptive signal processing, AM-PPO modulates GAE advantages, which encapsulate both reward and value function information, rather than raw rewards or returns alone. Furthermore, AM-PPO's modulation formula, involving scaling the absolute raw advantage by a controlled \(\tanh\)-based gate, and its specific controller for \(\alpha_A\), are distinct.

Another avenue for enhancing training stability involves regularizing the network parameters themselves. For instance, Spectral Normalized Actor-Critic (SNAC) \cite{Bawa2022SpectralNF} applies spectral normalization to both policy and value networks. This technique constrains the Lipschitz constant of the networks, thereby bounding the magnitude of policy gradients and critic TD updates, leading to smoother optimization and more stable training \cite{Bawa2022SpectralNF}. While SNAC and AM-PPO share the overarching goal of improving the conditioning of the learning process, their mechanisms are distinct. SNAC modifies the properties of the function approximators directly to ensure well-behaved gradients. In contrast, AM-PPO operates on the advantage signals themselves---inputs to the loss computation---by adaptively rescaling and transforming them based on their statistical properties and explicit targets, without directly regularizing the network weights. AM-PPO thus focuses on conditioning the data that informs the gradient, whereas SNAC focuses on constraining the functions that produce the gradients.

Other adaptive mechanisms in RL include dynamically adjusting exploration bonuses or entropy coefficients \cite{Chen2021ProximalPO}. While these methods also adapt aspects of the learning process, AM-PPO differentiates itself by directly manipulating the core advantage signal that drives both policy and value updates,
using a specific statistical feedback controller. The recently proposed Kalman Filter Enhanced GRPO \cite{Zhang2024KalmanFE} also utilizes adaptive advantage normalization in the context of GRPO, but its mechanism, leveraging Kalman filtering for state estimation and advantage processing, is substantially different from the controller-based, \(\tanh\)-gated modulation employed in AM-PPO.

In summary, while AM-PPO leverages principles of adaptive signal processing and non-linear transformations seen in other areas, its specific architecture—an adaptive controller targeting saturation of transformed GAE advantages, and the subsequent gated modulation of absolute raw advantages for both policy and value learning—represents a novel approach to conditioning learning signals in actor-critic reinforcement learning.

\section{Experimentation}
\label{sec:experimentation}

To empirically evaluate the performance and characteristics of AM-PPO, we conducted experiments on standard continuous control benchmarks from the OpenAI Gymnasium suite \cite{Gymnasium2023}. The primary goal was to understand not only the impact on aggregate reward but also how the advantage modulation mechanism influences intermediate learning signals, policy entropy, and its interaction with adaptive optimizers like DynAG \cite{DynAG2025}. All experiments were implemented using the CleanRL framework \cite{huang2022cleanrl}.

\subsection{Experimental Setup}
\begin{figure}[H]
    \centering
    \fbox{\includegraphics[width=1\linewidth]{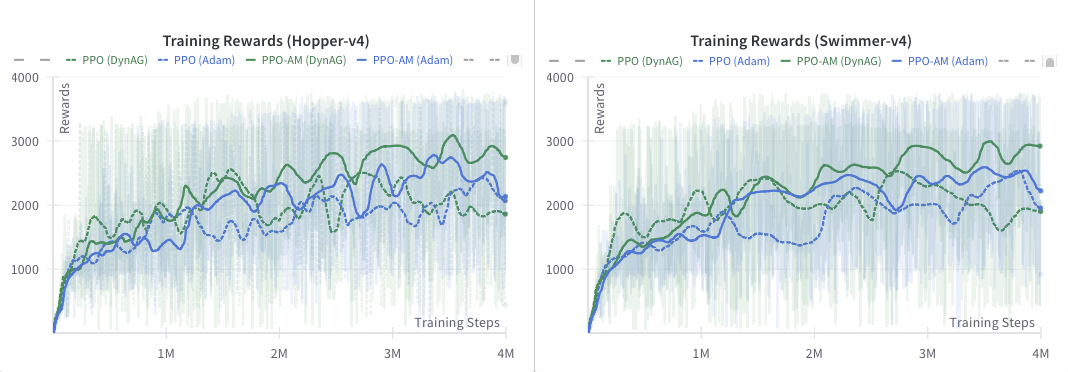}}
    \caption{Training Rewards for Gymnasium Environments}
    \label{fig:rewards_graph}
\end{figure}

We benchmarked AM-PPO against a standard PPO baseline. For both algorithms, core PPO hyperparameters (e.g., \(\gamma, \lambda, \epsilon_{\text{clip}}\)) were kept consistent, based on common settings for the chosen environments. We also decided to test against 2 optimizers: Adam \& Dynamic AlphaGrad. Key metrics recorded included episodic training rewards, policy loss (\(\mathcal{L}^{\pi}\)), value loss (\(\mathcal{L}^{V}\)), policy entropy (\(\mathcal{L}^S\)), the behavior of the adaptive scaling factor \(\alpha_{A,\text{ema}}\), the characteristics of modulated advantages, and, where applicable, the clipping fraction of the DynAG optimizer when used in conjunction with AM-PPO versus standard PPO. Figure \ref{fig:rewards_graph} showcases the rewards for all of the environments over training. All experiments were run with a set seed and achieved similar rewards, but PPO-AM constantly achieved a higher final reward. Furthermore, AM-PPO exhibited an interesting trend where the rewards were on a continual growth trajectory, whereas traditional PPO spiked early and was prone to plateaus. In the subsequent sections, we highlight the training dynamics of two of the environments. The full training dynamics of all of the environments tested are available in the appendix.

\subsection{Results on Swimmer-v4}
\begin{figure}[H]
    \centering
    \fbox{\includegraphics[width=1\linewidth]{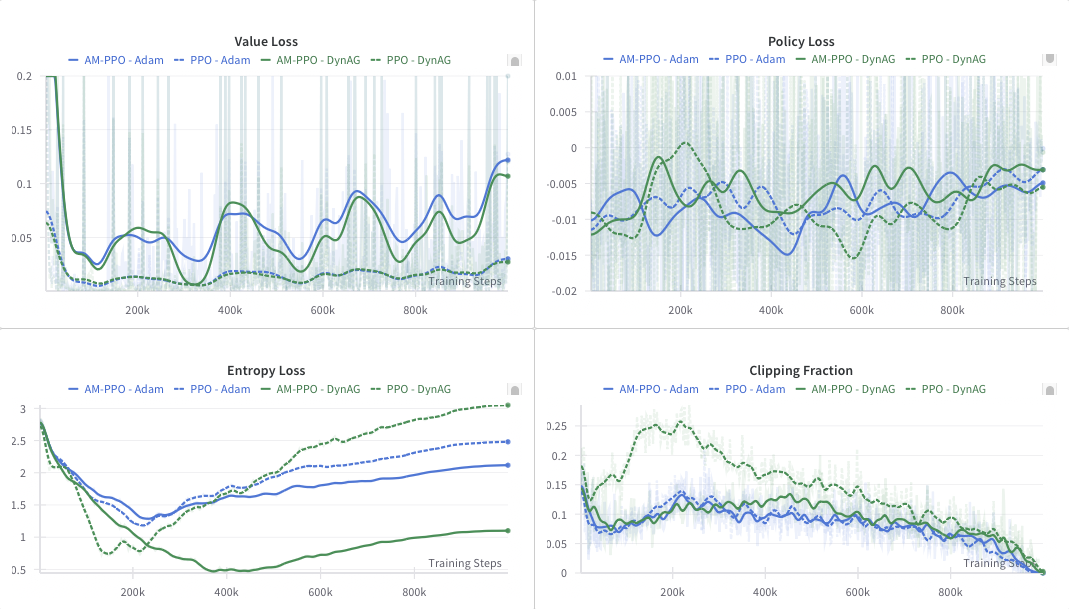}}
    \caption{Training Dynamics for Swimmer-v4}
    \label{fig:swimmer_entropy_dynag}
\end{figure}

In the Swimmer-v4 environment, apart from the variations in training rewards, significant differences emerged in other diagnostic metrics. Notably, AM-PPO exhibited consistently lower policy entropy throughout training (Figure~\ref{fig:swimmer_entropy_dynag}). This suggests that the modulated advantage signal might guide the policy towards more deterministic actions more rapidly or maintain a more focused exploration strategy.

Furthermore, when AM-PPO was used with the DynAG optimizer, the internal clipping fraction of DynAG was substantially suppressed compared to its behavior with standard PPO advantages (Figure~\ref{fig:swimmer_entropy_dynag}). This indicates that the modulated advantages presented to DynAG were perhaps scaled more appropriately or were less prone to causing large, clipped updates.

\begin{figure}[H]
    \centering
    \fbox{\includegraphics[width=1\linewidth]{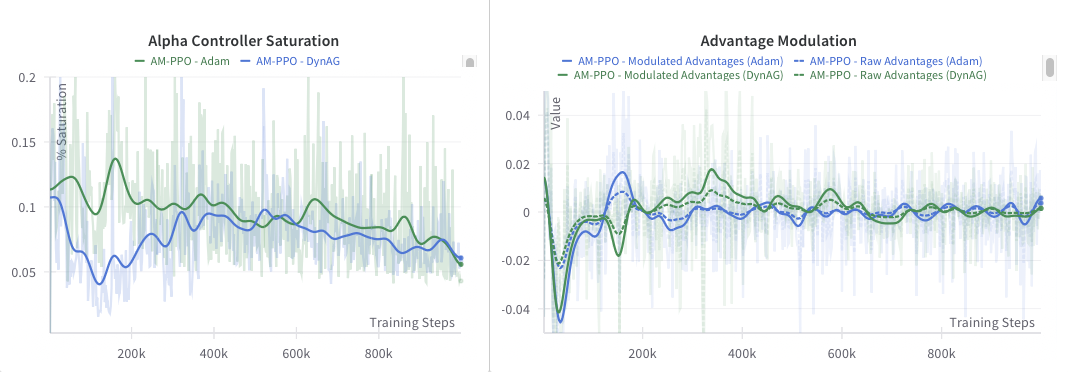}}
    \caption{Alpha Controller Dynamics for Swimmer-v4}
    \label{fig:swimmer_alpha_adv_mod}
\end{figure}

The adaptive controller for \(\alpha_{A,\text{ema}}\) operated as expected, adjusting the scaling factor based on advantage statistics and saturation targets, demonstrating its control feedback loop in action (Figure~\ref{fig:swimmer_alpha_adv_mod}). Multiplying by the magnitude of the raw advantages also proved to provide a natural isotropic decay over the course of the training (Figure~\ref{fig:swimmer_alpha_adv_mod}).

\subsection{Results on Hopper-v4}
\begin{figure}[H]
    \centering
    \fbox{\includegraphics[width=1\linewidth]{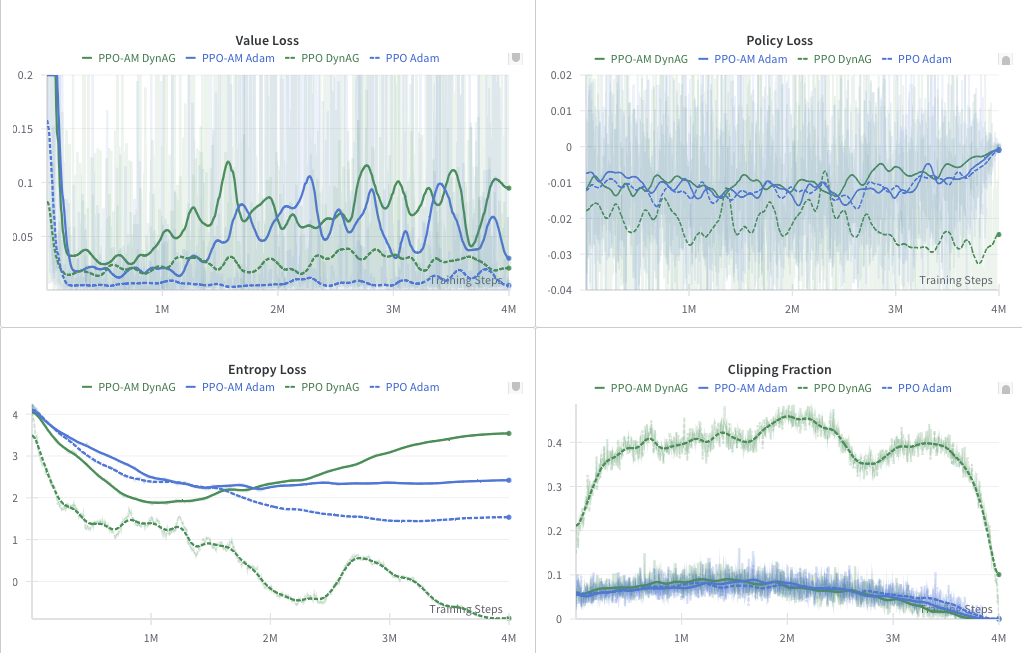}}
    \caption{Training Dynamics for Hopper-v4}
    \label{fig:hopper_entropy_dynag}
\end{figure}

In the Hopper-v4 environment, AM-PPO achieved higher asymptotic training rewards compared to the PPO baseline (Figure~\ref{fig:rewards_graph}). This performance improvement was, perhaps surprisingly, accompanied by a consistently higher value loss throughout much of the training, while the policy loss trajectories remained largely similar to the baseline (Figure~\ref{fig:hopper_entropy_dynag}).


A particularly interesting pattern emerged in the policy entropy for AM-PPO on Hopper-v4. While generally maintaining a higher entropy than the baseline for a significant portion of training, the entropy exhibited a "rebounding" behavior during periods where reward accumulation plateaued (Figure~\ref{fig:rewards_graph}). This suggests that the policy, under AM-PPO's influence, might adaptively increase its exploratory behavior when it detects stagnation in learning progress, potentially facilitating escape from local optima or challenging regions of the loss landscape. As with Swimmer, the DynAG clipping fraction was markedly suppressed when combined with AM-PPO (Figure~\ref{fig:hopper_entropy_dynag}).

\begin{figure} [H]
    \centering
    \fbox{\includegraphics[width=1\linewidth]{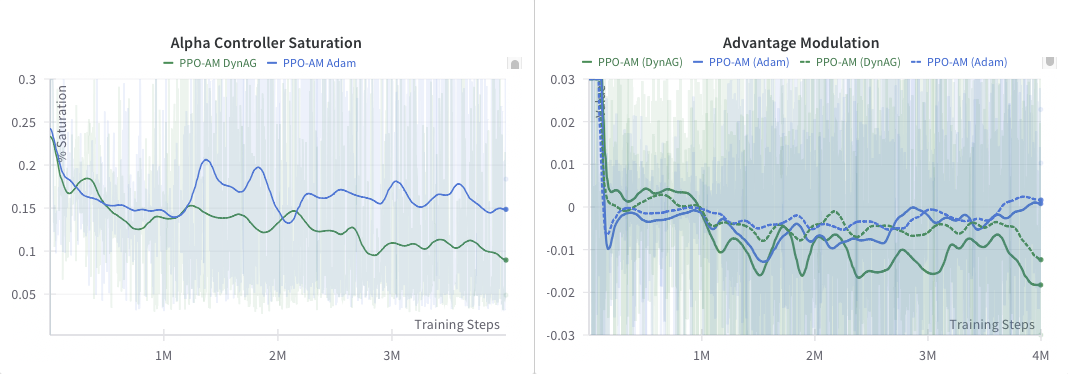}}
    \caption{Alpha Controller Dynamics for Hopper-v4}
\end{figure}

\subsection{Discussion of Experimental Findings}

Our empirical evaluation reveals that AM-PPO introduces beneficial modifications to the learning dynamics, leading to improved performance outcomes. Across the environments tested, AM-PPO not only consistently achieved higher final rewards (Figure~\ref{fig:rewards_graph}) but also demonstrated a notably different reward accumulation pattern. Specifically, AM-PPO facilitated a more sustained and progressive increase in rewards throughout training, exhibiting a continual growth trajectory. This contrasts with baseline PPO, which tended towards earlier reward spikes followed by more pronounced and prolonged plateaus. This characteristic of sustained learning progression suggests AM-PPO's potential for enhanced stability and performance in long-term training and its applicability to more complex scenarios where avoiding premature convergence is crucial. Complementing these reward dynamics, AM-PPO consistently led to a significant reduction in the internal clipping fraction of the DynAG optimizer across both environments. This indicates that the modulated advantages generated by AM-PPO are better conditioned, possessing a scale and distribution that mitigates the need for aggressive clipping by the optimizer, thereby potentially contributing to more stable and efficient policy updates.

We hypothesize that the AM-PPO modulation process, by introducing non-linear transformations and adaptive scaling, alters the effective loss landscape created by the policy. The "stretching" and reshaping of advantages could introduce more "ruggedness" or "spikiness" into plateaued regions of the gradient landscape. Such features, while potentially challenging for simpler gradient descent, might provide more informative signals for sophisticated optimizers to navigate and escape these areas. This contrasts with the potentially flatter, harder-to-escape plateaus that can characterize the optimization of large, complex models (analogous to observations in deep learning, e.g., comparing ResNet56 vs. ResNet110 loss landscapes \cite{Li2018VisualizingTL}).

A strong interpretation of this phenomenon can be characterized by the way in which modern optimizers traverse through intricate gradients - taking smaller steps where there is high magnitude - contrary to the popular SGD method. As models grow more complex, their loss domains increase in size and plateaus become longer and require more exploration. If we are able to terraform this plateau into a spikier gradient, we can be more careful to avoid falling down these plateaus and becoming trapped in an elongated local domain.

\begin{figure}[H]
    \centering
    \includegraphics[width=1\linewidth]{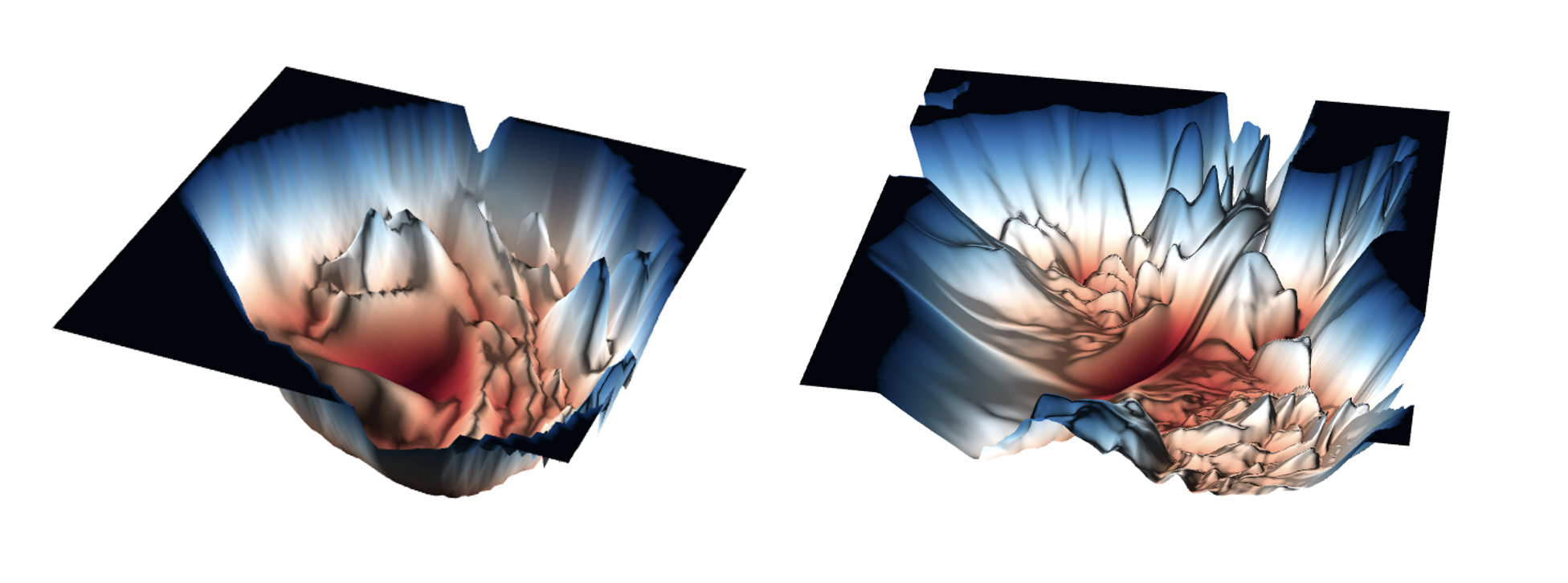}
    \caption{ResNet 110 (Left) vs ResNet 56 (Right) Loss Landscape}
\end{figure}

Apart from adjusting the gradient, the dynamic behavior of policy entropy, particularly the rebounding observed in Hopper during learning plateaus, is a salient finding. It suggests an implicit mechanism within AM-PPO that couples the perceived rate of learning progress (via advantage statistics) to the policy's exploratory drive. If the policy observes that its current gradient signals (derived from modulated advantages) are not leading to improvement, the system may naturally encourage broader exploration. Proof of this connection can be observed across both environments tested for only AM-PPO while traditional PPO only exhibited this connection for Swimmer-v4. Designing policies that explicitly link such entropy responses to gradient stagnation is a promising avenue for future work.

The observation of increased value loss in Hopper, despite achieving higher rewards, indicates that training the value function on targets derived from modulated advantages (Eq.~\ref{eq:gam_value_target_ppo_gam}) introduces a different dynamic. While \(A_{\text{mod}, j}\) may be beneficial for the policy gradient, its direct use as \(V_{\text{target}, j} - V_{\phi_{\text{old}}}(s_j)\) might pose a more challenging regression problem for \(V_\phi(s_j)\), or it could be that the value function is learning a different, perhaps more complex, representation. Further investigation into alternative value function learning targets within the AM-PPO framework is warranted.

In summary, while further research and optimization are needed across more than just MuJoCo style environments, AM-PPO proposes a novel perspective on advantage processing that continuously evolves past its initial experimentation. The results from the initial testing in this paper demonstrate its capacity to significantly alter learning dynamics, providing insights into the interplay between advantage signal conditioning, optimizer behavior, and policy exploration. The modulation technique itself appears to make the gradient landscape more informative for adaptive optimizers, potentially by introducing beneficial complexities that aid in escaping difficult optimization regions.

\section{Potential Extensions and Future Work}
\label{sec:agram_extension}

The introduction of AM-PPO and its adaptive advantage modulation mechanism opens several promising avenues for future research and extensions. The core concept of ``Alpha Modulation'' --- adaptively rescaling and transforming a group of signals based on their statistics and target properties --- can be explored in various contexts.

\subsection{AGRAM: Adaptive Group Relative Reward Modulation}
A primary direction is the development of Adaptive Group \textit{Reward} Alpha Modulation (AGRAM). Current AM-PPO modulates GAE advantages. AGRAM would instead apply a similar adaptive modulation directly to rewards sampled at a decision point. Specifically, at a given state \(s_t\), multiple actions \(a_i \sim \pi_\theta(\cdot|s_t)\) could be executed in the environment, yielding a set of observed rewards \(\{r_{t,i}\}\) and next states \(\{s_{t+1,i}\}\). The Alpha Modulation controller, as defined in Section~\ref{subsec:gam_modulation_mechanism} (but adapted for rewards, e.g., using statistics of \(\vb{r}_t\) instead of \(\vb{A}_{\text{raw}}\)), could then transform these raw rewards into modulated rewards \(\{\tilde{r}_{t,i}\}\).

To maintain a coherent trajectory for value function learning within typical actor-critic frameworks, the mean of these modulated rewards, \(\bar{\tilde{r}}_t = \mean{\{\tilde{r}_{t,i}\}}\), could be computed. The experience tuple \((s_t, a_k, r_{t,k}, s_{t+1,k})\) corresponding to the modulated reward \(\tilde{r}_{t,k}\) closest to this mean \(\bar{\tilde{r}}_t\) would then be selected and stored in the experience buffer. This approach aims to achieve two goals: first, it conditions the raw reward signal itself, potentially highlighting more salient reward information; second, it provides the value function \(V_\phi(s_t)\) with a target that reflects a more ``statistically representative'' outcome of the stochastic interactions at state \(s_t\), potentially stabilizing value estimation. Furthermore, the advantages derived from these AGRAM-selected experiences could subsequently undergo another layer of Alpha Modulation as in the current AM-PPO, creating a hierarchical signal conditioning pipeline. This differs from approaches like Hybrid GRPO \cite{Sane2025HybridGRPO} which might incorporate all sampled points into the buffer (while conserving the value function) after a similar \(\tanh\) transformation, a strategy particularly beneficial under severe data scarcity. AGRAM, in this initial conception, focuses on refining the signal for a single representative trajectory, though exploring the use of all modulated samples as extracted data remains a viable research path.

\subsection{Value Function-Free Alpha Modulation}
Inspired by policy optimization methods that operate without a learned value function (e.g., certain variants of REINFORCE \cite{zhang2020sampleefficientreinforcementlearning} or approaches like the one suggested by GRPO \cite{Shao2024GRPO} which may eschew traditional value networks), AM-PPO's Alpha Modulation could be investigated in such contexts. If raw Monte Carlo returns \(G_t\) are used in lieu of GAE advantages \(A^{\text{raw}}_t\), Alpha Modulation could be directly applied to these returns. This would represent a direct, adaptive, non-linear transformation of the empirical sum of future rewards, potentially stabilizing policy gradients derived from high-variance Monte Carlo estimates. This would also directly mitigate the higher variance and value loss observed in the experimentation. The framing of AM-PPO as a ``higher-order adjustment of the integral'' (the return) would become particularly salient in this setting.

\subsection{Alpha Modulation for Q-Learning and Off-Policy Algorithms}
The principle of Alpha Modulation is not inherently limited to PPO or on-policy learning. Its application to Q-learning frameworks warrants investigation. For instance, Alpha Modulation could be applied to the stream of Temporal Difference (TD) errors, \(\delta_t = (r_t + \gamma \max_{a'} Q_{\text{target}}(s_{t+1}, a')) - Q(s_t, a_t)\), before they are used to update \(Q(s_t, a_t)\). By adaptively rescaling and transforming these TD errors based on their batch statistics (e.g., norm, standard deviation, and saturation targets analogous to those in AM-PPO), the learning signal for the Q-function could be better conditioned.

Off-policy algorithms, which often leverage large replay buffers containing diverse and potentially temporally uncorrelated experiences, might particularly benefit. Alpha Modulation could help normalize the learning signals derived from these varied transitions, dynamically adjusting their influence based on their statistical properties within a sampled batch. This could potentially mitigate issues arising from stale data or high variance in off-policy targets, thereby improving the stability and efficiency of Q-learning and related actor-critic methods like DDPG or SAC.

\section{Final Remarks}
\label{sec:final_remarks}

This paper introduced AM-PPO, a novel extension to Proximal Policy Optimization that incorporates an adaptive mechanism, termed Alpha Modulation, to condition advantage estimates. Our experiments demonstrate that this technique can lead to more sustained reward growth, influence policy entropy, and beneficially interact with adaptive optimizers by reshaping the learning signals.

The core contribution we wish to emphasize extends beyond PPO itself: Alpha Modulation represents a potentially generalizable mathematical framework for adaptively transforming groups of data points. By leveraging statistics of a data collective (like norms and standard deviations) in conjunction with non-linear transformations (such as the \(\tanh\) function driven by a saturation target), it offers a new way to achieve a form of differentiable, relative conditioning. The underlying theory suggests that relationships within a set of signals are rarely purely linear, and Alpha Modulation provides a mechanism to capture and exploit these non-linear dependencies to refine learning signals. While this work presents its initial application and theory in the context of advantage estimation, the full implications and theoretical underpinnings of Alpha Modulation as a data conditioning technique warrant significant further research.\\

Special thanks to Angus McLean (Collins Aerospace) for insights, discussions, \& mentorship on the development of DynAG \& AM-PPO. Thank you to Lambda Labs for the GPU access to perform development and testing. Thank you also to Weights \& Biases for their integration to observe training results and dynamics, and for the organization of runs and projects.

All work presented in this paper was conducted in the spirit of open-source contribution and research. The broader research community is encouraged to build upon, critique, and extend these findings. Further exploration and collaborative efforts are welcomed to fully understand and harness the potential of Alpha Modulation in reinforcement learning and potentially other domains of machine learning.

\newpage
\appendix

\newpage
\appendix

\section*{Appendix A: Limited Ablation Studies}

\subsection*{Ant-v4 Training Dynamics Ablation Study}
\label{app:ant_dynamics}

A notable aspect of the primary experiments presented is that configurations of PPO utilizing the DynAG optimizer \cite{DynAG2025} were implemented with momentum, a common practice for this optimizer. Conversely, AM-PPO, when paired with DynAG, achieved strong performance without explicit optimizer momentum. This observation prompted an inquiry into the interplay between AM-PPO's intrinsic signal conditioning and the role of optimizer momentum. The theoretical premise is that AM-PPO's direct modulation of advantage signals may establish a more effectively conditioned learning landscape, potentially diminishing the necessity or benefit of momentum's gradient accumulation mechanisms. AM-PPO's adaptive scaling, responsive to the evolving distribution of advantages, and the observed entropy rebounding phenomena (indicative of self-regulating exploration), might offer a more direct means of navigating the optimization landscape than conventional momentum. This suggests that for certain reinforcement learning contexts, enhancing signal quality at its source could be more impactful than relying on general-purpose optimization heuristics largely derived from supervised learning.

\begin{figure}[H]
    \centering
    \fbox{\includegraphics[width=0.9\linewidth]{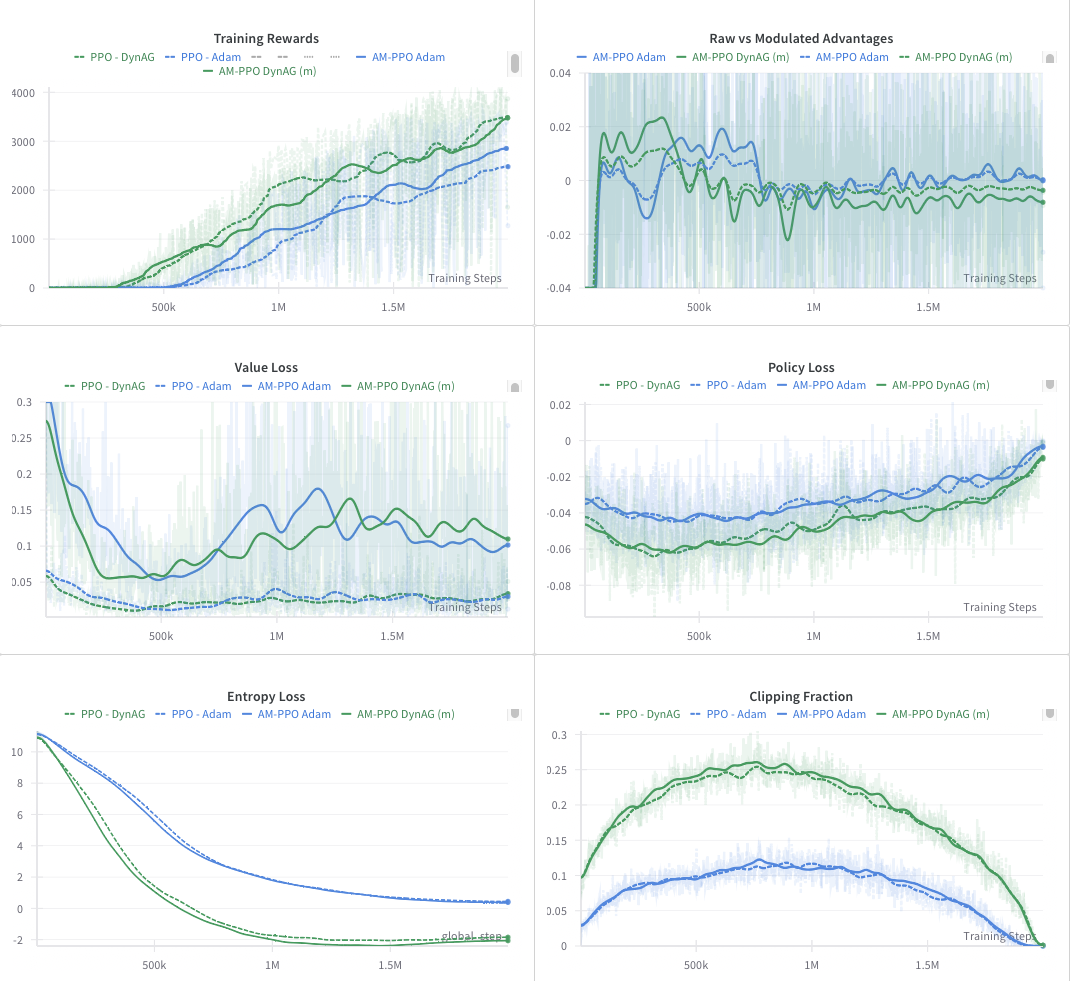}} 
    \caption{Ant-v4 Training Dynamics}
    \label{fig:ant_fig_ablation}
\end{figure}

To investigate this hypothesis, a targeted ablation study was conducted. This study focused on observing AM-PPO with the DynAG optimizer when momentum was explicitly re-introduced. Additionally, comparative data was collected for AM-PPO versus PPO using the Adam optimizer. For this ablation, learning rates were specifically tuned for each optimizer-algorithm pairing, departing from the consistent learning rate approach of the initial broader experiments. The results of this Ant-v4 ablation are presented in Figure~\ref{fig:ant_fig_ablation}.

The findings from this ablation lend support to the hypothesis that AM-PPO's advantage modulation, operating without optimizer momentum, contributes significantly to its convergence characteristics. Specifically, when momentum was added to AM-PPO DynAG, while reward trajectories remained competitive within the 2 million step observation window, the notable suppression of the DynAG clipping fraction observed in momentum-free AM-PPO DynAG configurations (detailed in the main paper's experiments) was less evident. This outcome, observed even with tuned learning rates, raises pertinent questions: Does Alpha Modulation inherently provide greater tolerance in learning rate selection? Furthermore, does this suggest that optimization strategies rooted in RL-specific signal conditioning, such as advantage modulation, hold greater promise than direct adoption of techniques from supervised learning paradigms? These questions delineate critical avenues for future research.

\subsection*{Humanoid-v4 Training Dynamics Ablation Study}
\label{app:humanoid_dynamics}

\begin{figure}[H]
    \centering
    \fbox{\includegraphics[width=0.9\linewidth]{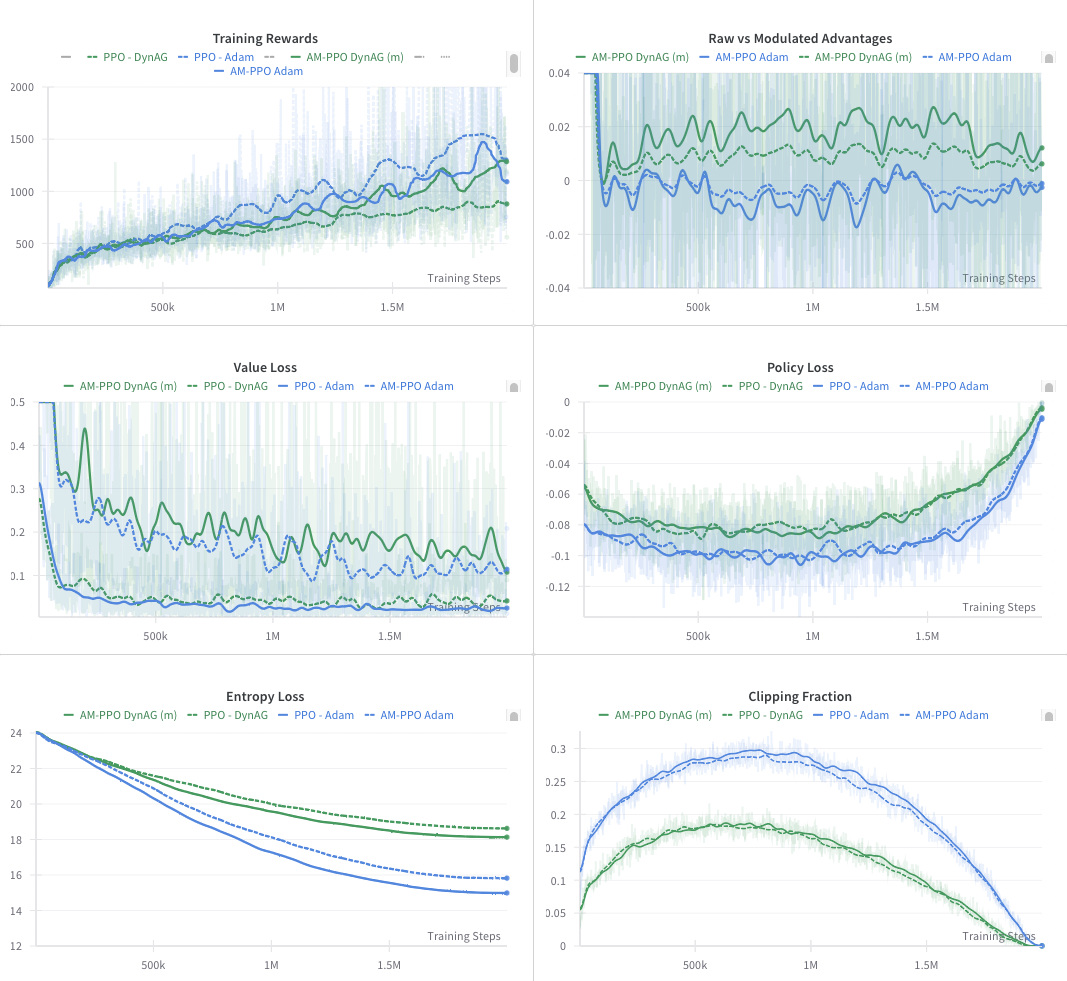}} 
    \caption{Humanoid-v4 Training Dynamics}
    \label{fig:humanoid_fig_ablation}
\end{figure}

The training dynamics of the Humanoid-v4 experiment, under the same ablation conditions as Ant-v4 (tuned learning rates, specific investigation of momentum's role with DynAG \cite{DynAG2025}), are illustrated in Figure~\ref{fig:humanoid_fig_ablation}.

In the Humanoid-v4 environment, AM-PPO paired with momentum DynAG optimizer \cite{DynAG2025} demonstrated superior reward accumulation compared to the standard PPO with a momentum-equipped DynAG optimizer by the 2 million step mark. Conversely, AM-PPO with Adam exhibited slightly lower rewards than PPO with Adam within this timeframe, although both configurations displayed continued learning potential. It remains crucial to acknowledge the 2 million step limitation of these experiments, which may not fully capture asymptotic performance characteristics or allow for the convergence of all algorithmic variants.

Consistent with the Ant-v4 ablation, the most salient finding for Humanoid-v4 pertains to the interaction with DynAG \cite{DynAG2025}. The pronounced suppression of the DynAG clipping fraction and the beneficial entropy patterns, characteristic of momentum-free AM-PPO DynAG (as observed in primary experiments), were less apparent when considering the hypothetical inclusion of momentum, even with individually tuned learning rates. This replication of findings across disparate environments reinforces the hypothesis regarding the fundamental impact of advantage modulation on the optimization landscape, potentially obviating or altering the traditional role of optimizer momentum.

\section*{Appendix B: Future Experimentation Roadmap}
\label{app:future_experimentation}

The empirical results presented herein, while promising, were circumscribed by available computational resources. Consequently, a comprehensive exploration of AM-PPO's capabilities warrants an extended experimental program. Table~\ref{tab:future_experiments_symbolic} outlines key directions for future investigation.

\begin{table}[H]
    \centering
    \caption{Proposed Future Research \& Ablation Studies for AM-PPO.}
    \label{tab:future_experiments_symbolic}
    \setlength{\tabcolsep}{3pt} 
    \renewcommand{\arraystretch}{1.2} 
    \begin{tabular}{|l|l|p{0.53\textwidth}|} 
        \hline
        \textbf{Category} & \textbf{Focus / Variable} & \textbf{Objective / Investigation} \\
        \hline
        Scalability & Training Duration & Assess asymptotic performance \& stability (\(>\)5M steps). \\
        \hline
        Robustness & Hyperparameters (\(\kappa, p_{\star}, \eta, \rho\)...) & Sensitivity analysis; establish optimal ranges/tuning. \\
        \hline
        Optimization & Optimizers (SGD, RMSProp...) & Interaction with/without momentum; comparative efficacy. \\
        & Learning Rates & Tolerance; interaction with AM-PPO's conditioning. \\
        \hline
        Generalization & Environment Suite & Diverse tasks (discrete, sparse, pixels, e.g., Atari, Procgen). \\
        & Network Architecture & Impact of model size/complexity on AM-PPO benefits. \\
        \hline
        Mechanism & Controller Design & Alternative statistics, EMA, target metrics for \(\alpha_A\). \\
        & Modulation Function & Explore alternatives to \(\tanh\). \\
        \hline
        Synergy & Combined Techniques & Interaction with PER, intrinsic motivation, multi-step learning. \\
        \hline
    \end{tabular}
\end{table}

Systematic execution of these studies would furnish more conclusive evidence regarding the generalizability, robustness, and optimal deployment of the advantage modulation paradigm. The current findings strongly motivate such extended research to fully delineate the contributions of AM-PPO to reinforcement learning.

These experiments would provide more comprehensive evidence about the generalizability of the advantage modulation approach and its fundamental importance in reinforcement learning optimization. While current results are promising, the full potential and robustness of AM-PPO may only be realized through more extensive testing across diverse conditions and longer training horizons.

\section*{Appendix C: Hyperparameters and Resource Note}
\label{app:hyperparams_and_resource}

This research was conducted by a solo researcher with computational resources primarily consisting of a single GH200 GPU rented from Lambda Labs. While these resources necessitated careful planning and limited the scale of some explorations, the consistent patterns observed provide compelling preliminary evidence for the efficacy of the advantage modulation approach. The code for AM-PPO and the experiments will be made available.\footnote{Experiment Repository - \url{https://github.com/Soham4001A/CleanRLfork}}

The following tables detail the hyperparameters used for the PPO baseline and the AM-PPO specific modulation mechanism across the primary experiments.

\subsection*{PPO Core Hyperparameters}
Table~\ref{tab:ppo_core_hyperparams_centered} lists the standard PPO hyperparameters used as a baseline and within AM-PPO.

\begin{table}[H]
    \centering
    \caption{PPO Core Hyperparameters for All Experiments.}
    \label{tab:ppo_core_hyperparams_centered}
    \begin{tabular}{@{}lc@{}} 
        \toprule
        \textbf{Parameter}               & \textbf{Value/Description}                                     \\ \midrule
        Total Timesteps         & 1,000,000                                             \\
        Number of Environments (\(N_{\text{envs}}\)) & 1                                                       \\
        Rollout Buffer Steps (\(N_{\text{steps}}\)) & 2048                                                  \\
        Anneal Learning Rate    & True                                                  \\
        Discount Factor (\(\gamma\)) & 0.99                                                  \\
        GAE Lambda (\(\lambda\))    & 0.95                                                  \\
        Number of Minibatches   & 32                                                    \\
        PPO Update Epochs (\(K\)) & 10                                                    \\
        Clipping Coefficient (\(\epsilon_{\text{clip}}\)) & 0.2                                                   \\
        Clip Value Loss         & True                                                  \\
        Entropy Coefficient (\(\beta_{\text{ent}}\)) & 0.0                                                   \\
        Value Function Coeff. (\(c_V\)) & 0.5                                                   \\
        Max Gradient Norm       & 0.5                                                   \\
        Target KL Divergence    & None                                                  \\ \bottomrule
    \end{tabular}
\end{table}

\subsection*{AM-PPO Advantage Modulation Hyperparameters}
Table~\ref{tab:am_ppo_specific_hyperparams_centered} details the hyperparameters specific to the AM-PPO advantage modulation controller. These correspond to the parameters described in Section~\ref{subsec:gam_modulation_mechanism} and Table~\ref{tab:ppogam_hyperparams} in the main text.

\begin{table}[H]
    \centering
    \caption{AM-PPO Specific Advantage Modulation Hyperparameters.}
    \label{tab:am_ppo_specific_hyperparams_centered}
    \begin{tabular}{@{}lc@{}} 
        \toprule
        \textbf{Symbol (from text)} & \textbf{Value} \\ \midrule
        \(\tau_A\) & 1.25 \\
        \(p_{\star,A}\) & 0.10 \\
        \(\kappa_{\text{shared}}\) & 2.0 \\
        \(\eta_A\) & 0.3 \\
        \(\rho_A\) & 0.1 \\
        \(\epsilon_A\) & \(1 \times 10^{-5}\) \\
        \(\alpha_{\min,A}\) & \(1 \times 10^{-12}\) \\
        \(\alpha_{\max,A}\) & \(1 \times 10^{12}\) \\
        \(\rho_{\text{sat},A}\) & 0.98 \\
        \(\alpha_{A,\text{ema}}^{(0)}\) & 1.0 \\
        \(s_{\text{prev},A,\text{ema}}^{(0)}\) & 0.10 \\
        \bottomrule
    \end{tabular}
\end{table}

The Alpha Modulation technique introduced in this paper represents not just an improvement to PPO, but potentially a new paradigm for thinking about learning signal conditioning in reinforcement learning. By focusing on the adaptive and relative processing of the quality and characteristics of the signals themselves, rather than just the optimization process that consumes them, this work opens numerous avenues for future research and application.

\end{document}